# GROUP EMOTION RECOGNITION USING MACHINE LEARNING

**Third Year Project Report**

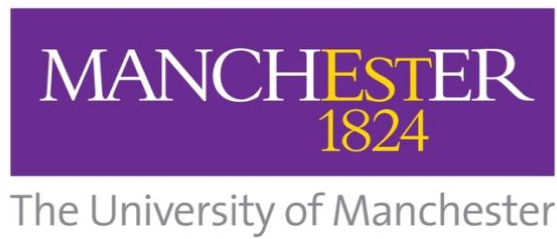

**Samanyou Garg**

Supervisor: Prof. Angelo Cangelosi

School of Computer Science
The University of Manchester
United Kingdom

Submitted in partial fulfilment for the degree of
*Bachelor of Science in Computer Science
with Industrial Experience*

Submitted April 2019

# Abstract

Automatic facial emotion recognition is a challenging task that has gained significant scientific interest over the past few years but the problem of emotion recognition for a group of people has been less extensively studied. However, it is slowly gaining popularity due to the massive amount of data available on social networking sites containing images of groups of people participating in various social events.

Group emotion recognition is a challenging problem due to obstructions like head and body pose variations, occlusions, variable lighting conditions, variance of actors, varied indoor and outdoor settings and image quality.

The objective of this task is to classify a group's perceived emotion as Positive, Neutral or Negative.

In this report, I describe my solution which is a hybrid machine learning system that incorporates deep neural networks and Bayesian classifiers.

- Deep Convolutional Neural Networks (CNNs) work from bottom to top, analysing facial expressions expressed by individual faces extracted from the image.
- The Bayesian network works from top to bottom, inferring the global emotion for the image, by integrating the visual features of the contents of the image obtained through a scene descriptor.
- In the final pipeline, the group emotion category predicted by an ensemble of CNNs in the bottom-up module is passed as input to the Bayesian Network in the top-down module and an overall prediction for the image is obtained.

Experimental results show that the stated system achieves 65.27% accuracy on the validation set which is in line with state-of-the-art results.

As an outcome of this project, a Progressive Web Application and an accompanying Android app with a simple and intuitive user interface are presented, allowing users to test out the system with their own pictures.

# Acknowledgements

I would first like to thank my supervisor, Prof. Angelo Cangelosi, for his constant guidance, feedback, and support throughout the final year. I have been fortunate enough to have a supervisor who believed in me and encouraged me to give my best to the project. I would like to express my gratitude to Dr. Lucas Cordeiro for providing valuable feedback in his role as my second marker. I would also like to thank Luca Surace for his help throughout the project.

I am indebted to my parents and my sister for their continued support and encouragement. They are the most important people in my life, and I am grateful to them for their help.

I would also like to thank all the members of staff at the University of Manchester for their knowledge and assistance.

I acknowledge the contribution of everyone who tested my app; this has been incredibly helpful for my project.

Finally, I would like to thank the organizers of the EmotiW challenge for generously providing me with the GAF 3.0 dataset [1] for this project. It has certainly made this task a lot easier.

# Contents

# List of Figures

# List of Tables

# Foreword

This report assumes prior familiarity with the following topics:

1. **Neural Networks,** including the following:

    a. Perceptrons
    b. The notion of input and hidden layers
    c. Loss functions
    d. (Stochastic) Gradient Descent algorithm
    e. Learning Rate
    f. Regularization
    g. Backpropagation
    h. Activation functions (sigmoid, tanh, SoftMax)
    i. Hyperparameters

2. **Bayesian Networks**

3. **Basic Image Processing techniques**

4. **Linear Algebra**

# Chapter 1 - Introduction

Automatic emotion recognition is a challenging task that has gained significant scientific interest in recent years due to its applications in crowd analytics, social media, marketing [2], event detection and summarization, public safety [3], human-computer interaction [4], digital security surveillance, street analytics, and image retrieval [5]. However, the problem of emotion recognition for a group of people has been less extensively studied, but it is slowly gaining popularity due to the massive amount of data available on social networking sites containing images of groups of people participating in social events.

Usually, emotion recognition is accomplished through the analysis of a set of facial muscle movements that correspond to a displayed emotion (also known as action units). After detecting and separating the subject face, feature extraction can be applied to predict the contribution of each action unit. However, this approach becomes impractical in unregulated environments where factors like head and body pose variations, occlusions, variable lighting conditions, variance of actors, varied indoor and outdoor settings and image quality affect the recognition of emotions.

## 1.1 Aims and Objectives

This project aims to develop an automatic system that can reliably classify a group's perceived emotion as Positive, Neutral or Negative. Furthermore, the system should work appropriately on any device irrespective of the processing power available on the user's device.

The following objectives help to achieve the overall goal of the project:

- Conduct research on different approaches for group emotion recognition in the wild.
- Design and implement one approach and perform experiments on various aspects of the chosen approach to improve it further.
- Carry out performance analysis of the different methods and models.
- Develop a web app and an accompanying Android app, allowing users to upload an image for emotion recognition from their PC or taken immediately from the camera of their smartphone/tablet. Also, ensure that the user interface for both the desktop and mobile applications is easy to use.

## 1.2 Proposed Solution

In this paper, we present a solution that combines both local and global information. Our solution builds on the model by Surace et al. [6] and extends it further with additional and more refined machine learning methods and experiments.

- The bottom-up module detects and extracts individual faces present in the image and passes them as input to an ensemble of pre-trained Deep Convolutional Neural Networks (CNNs).
- Simultaneously, the top-down module detects the labels associated with the scene and passes them as input to a Bayesian Network (BN) which predicts the probabilities of each class.
- In the final pipeline, the group emotion category predicted by the bottom-up module is passed as input to the Bayesian Network in the top-down module and an overall prediction for the image is obtained.

We tested the system on the GAF 3.0 dataset [1] (released for the Emotion Recognition in the Wild Challenge 2018 (EmotiW)) and obtained an accuracy of 65.27% on the validation set which is similar to the baseline results [1] obtained by the organisers of the competition who used a very deep Inception v3 architecture.

As an outcome of this project, a Progressive Web Application and an accompanying Android app with a simple and intuitive user interface are presented, allowing users to test out the system with their own images.

## 1.3 Report Structure

The report consists of 9 chapters:

- **Chapter 1: Introduction -** Outline the problem, describe the aims and objectives of the project and introduce the proposed solution.

- **Chapter 2: Related Work -** Review previous research papers on group emotion recognition.

- **Chapter 3: Dataset -** Exploratory data analysis on the GAF 3.0 dataset.

- **Chapter 4: Method -** Describe in detail the machine learning based approach that was employed to tackle the problem.

- **Chapter 5: Experimental Design -** Outline the series of experiments that were carried out as well as the methodology that was used to improve the performance of the proposed method.

- **Chapter 6: Evaluation -** Describe the evaluation method used to test the entire system as well as present the results.

- **Chapter 7: Tools and Technologies -** Describe the tools and technologies used to implement the proposed approach.

- **Chapter 8: Deliverables -** Present the web and the android applications developed as an outcome of this project.

- **Chapter 9: Conclusion -** Summarise the achievements of the project, the challenges encountered and the improvements for future work.

# Chapter 2 – Related Work

Quite a few research papers having "group emotion recognition" as the object of study have been written over the past 3-4 years. For example, in [1], a deep CNN model (Inception V3 network followed by three fully connected layers) was used to accomplish the task. In [7], Li proposed the use of a convolutional neural network (CNN) model to extract discriminative facial features and a long short-term memory (LSTM) to selectively learn the important features for the group-level happiness prediction task. The authors used a bottom-up approach where they extracted geometric features from 49 facial points and then used these features to train a model using Partial Least Squares regression in [8]. In [9], Sun proposed a combination of LSTM and GEM models for group Emotion recognition in the wild. In [10], a combination of Reisz-Based volume local binary pattern and continuous conditional random fields model was used to accomplish the task. The Reisz-Based volume local binary pattern considers neighbouring changes in the spatial domain of a face as well as different Riesz faces whereas the conditional random fields consider both global and local attributes. In [11], the authors explored face, body and context features for the inference of group emotion using arousal and valence emotion annotations. In [12], the authors used a combination of two types of CNNs, namely individual facial emotion CNNs and global image based CNNs to recognize group-level emotion in images.

# Chapter 3 - Dataset

The dataset that is being used for this project is the "Group Affect Database 3.0" [1]. It has been generously provided by the organisers of a competition called the EmotiW challenge.

The dataset contains "in the wild" photos of groups of people in various social environments and is split into training and validation sets of 9815 and 4346 images (of varying sizes) separately.

Both the training and validation sets are divided into 3 classes – Positive, Neutral and Negative corresponding to the 3 emotion categories. Figure 3.1 depicts one example of each category.

Numerous other research papers that were published as part of the EmotiW challenge used this dataset for the evaluation of their respective methods allowing us to compare the accuracy across different architectures easily.

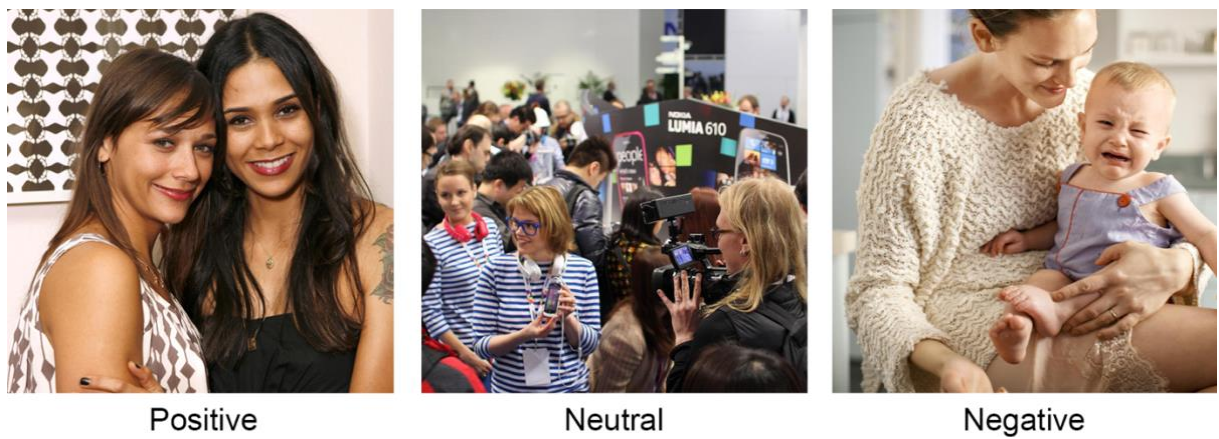

Figure 3.1 - Example images from the "Group Affect Database 3.0" demonstrating the 3 emotion categories.

The dataset is challenging due to obstructions like head and body pose variations, occlusions, variable lighting conditions, variance of actors, varied indoor and outdoor settings and image quality. Also, the training set is imbalanced, i.e. the number of images in the training set varies per class (see Figure 3.2). As a result, we can expect a convolutional neural network (CNN) to generalise poorly on the negative emotion as it has a lower number of training samples.

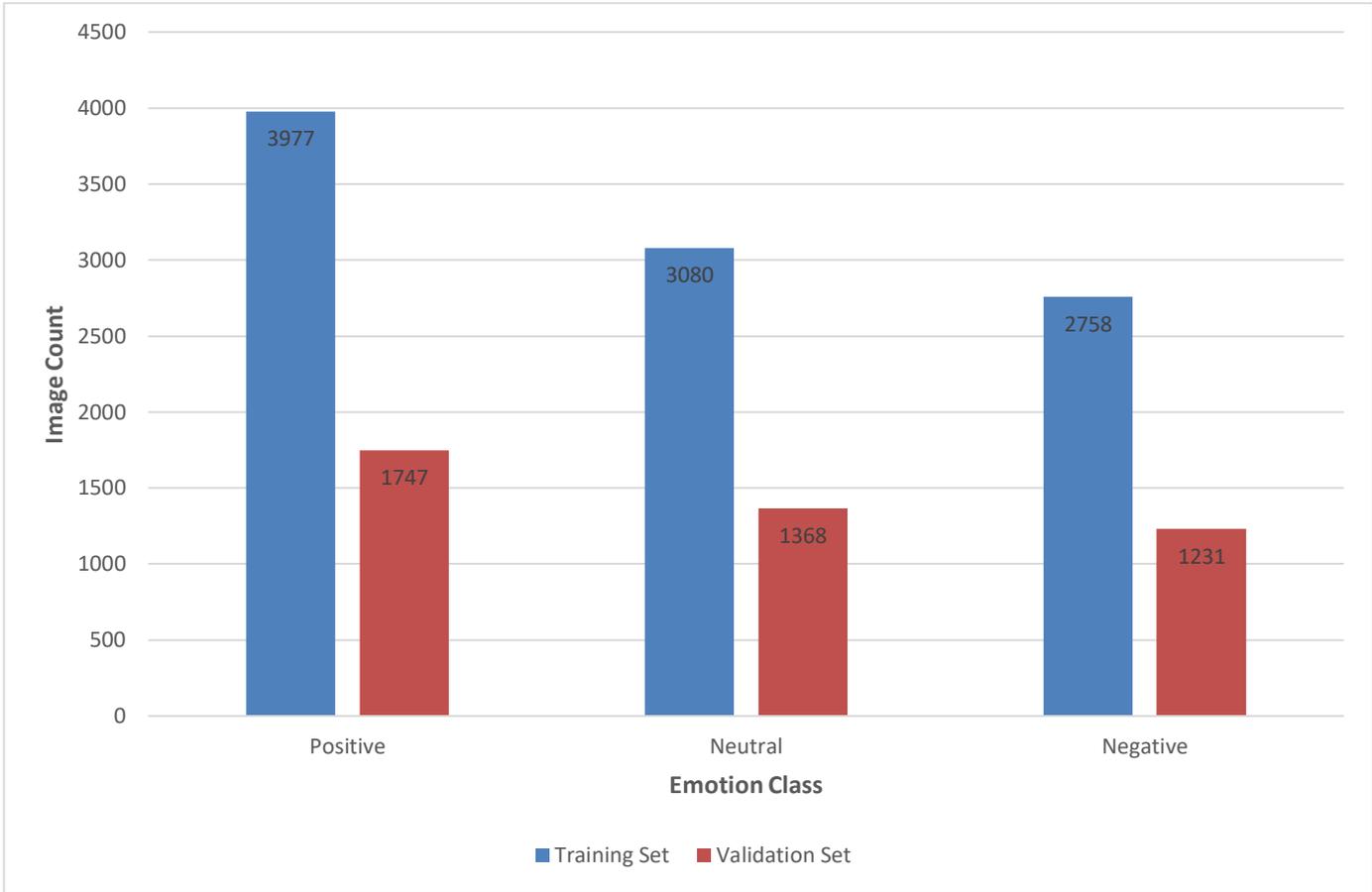

Figure 3.2 - Chart showing the number of images in the GAF 3.0 training and validation sets for each class.

# Chapter 4 - Method

Before starting the development of this project, a significant amount of research was carried out on which approach to follow for this task. As a result of our analysis, we decided to use a method that takes into account both local and global information.

Our approach for this project is inspired by the research paper "Emotion Recognition in the Wild using Deep Neural Networks and Bayesian Classifiers" [6]. It is a pipeline-based approach which integrates two modules (that work in parallel): bottom-up and top-down modules, based on the idea that the emotion of a group of people can be deduced using both bottom-up [13] and top-down [14] approaches.

The bottom-up module detects and extracts individual faces present in the image and passes them as input to an ensemble of pre-trained Deep Convolutional Neural Networks (CNNs).

Simultaneously, the top-down module detects the labels associated with the scene (background of the image) and passes them as input to a Bayesian Network (BN) which predicts the posterior probabilities of each class.

In the final pipeline, the group emotion category predicted by the bottom-up module is passed as input to the Bayesian Network in the top-down module and an overall group emotion prediction for the image is obtained. An overview of the full pipeline is shown in Figure 4.1.

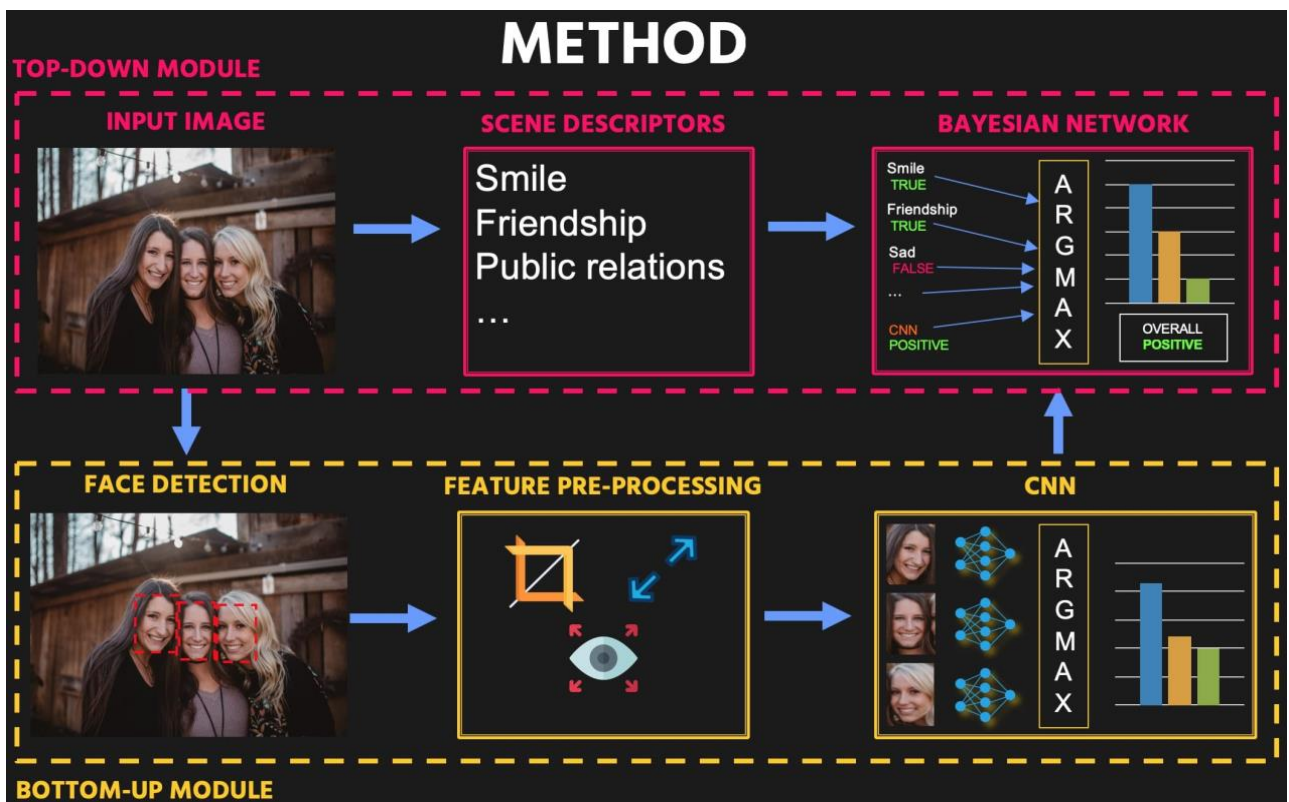

Figure 4.1 – Overview of the full pipeline.

The **bottom-up module** consists of the following four steps -
1. *Face detection*
2. *Feature pre-processing*
3. *Individual facial emotion CNN forward pass*
4. *Averaging of predictions*

The first step of the bottom-up module is face detection which was achieved using a combination of HoG face detection algorithm [15] and Multi-task Cascaded Convolutional Networks (MTCNN) algorithm [16]. This step returns a list of coordinates of all the faces detected in the image. In the second step, the features obtained from the first step are cropped, scaled, normalized and aligned. In the third step, the pre-processed features are fed into a pre-trained CNN model (trained on isolated facial images [See Section 5.3 for details]) and an array containing the individual emotion predictions for each face is obtained. Finally, an average of all the array elements is computed in order to get the overall group emotion prediction.

The **top-down module** consists of the following three steps -
1. *Acquiring the scene descriptors*
2. *Setting evidence in the Bayesian Network*
3. *Estimating the posterior probabilities of the Bayesian Network*

The first step of the top-down module is the acquiring of scene descriptors. Here, a label detection API [17] is used to estimate the content of the image. It returns a list of context-specific (e.g., party, vacation, picnic, etc.) and group-specific (e.g., friends, school, military, etc.) labels as output. In the second step, the scene descriptors obtained in the previous step are used to set evidence in the Bayesian network. Finally, in the third step, the belief propagation algorithm [18] is used to perform inference on the Bayesian network and the probabilities of each emotion in the root node are obtained.

## 4.1 Bottom-up Module

### 4.1.1 Face Detection

Several state-of-the-art facial detection algorithms and APIs were implemented and tested in order to identify the algorithm that could detect the maximum number of individual faces in each group image [detailed explanation of the facial detection algorithms tried can be found in Section 5.1].

After careful experimentation, a combination of HoG face detection algorithm [15] and Multi-task Cascaded Convolutional Networks (MTCNN) algorithm [16] was chosen as this combination was able to achieve the fastest performance and the highest accuracy in facial detection task on "in the wild" images. It returns a list containing coordinates of isolated human faces detected in the group image as output.

## 4.1.2 Feature pre-processing

Feature pre-processing techniques are the transformations performed on raw image data before feeding it to a deep learning algorithm (CNN in this case). Various advanced techniques were explored in order to improve the classification performance of the CNN as well as to speed up the training process [detailed explanation of the techniques tried can be found in Section 5.2] and cropping, scaling, normalization and face alignment were chosen as a result.

## 4.1.3 CNN forward pass

A pre-trained convolutional neural network (CNN) was used to predict the emotions of isolated human faces. CNNs are a type of deep neural networks that are most commonly used for image classification purposes. They can take an image as input, assign importance to various features of the image and then differentiate one image from the other. CNNs use a variation of multilayer perceptrons designed to require minimal pre-processing. They have been some of the most prominent innovations in the field of computer vision.

### 4.1.3.1 Individual Facial Emotion CNN Architecture

Our CNN accepts 64 x 64 x 3 array of pixel values (i.e., images of size 64x64 with 3 channels (RGB)) as input. Several state-of-the-art CNN models with very diverse network architectures have been implemented and tested for this project, starting from shallow networks with few layers to deeper models with a large number of layers.

Table 4.1 shows some of the CNN models with their corresponding number of layers, descriptions, and accuracies (on the validation set of isolated faces dataset). [*Network architectures of some of the best performing models from this table can be found in Appendix A.*] It can be observed from Table 4.1 that our earlier models employed 3-5 convolutional layers and performed significantly worse than the later models which consisted of 7+ convolutional layers. Increasing the number of convolutional layers to seven notably boosted the classification accuracy. However, increasing the number of layers beyond that did not prove to be very worthwhile.

As a result, Model 5 was chosen to be the default model for the rest of the experiments as it performed adequately and produced low validation loss. It consists of 7 convolutional layers, 3 max-pooling layers, and 3 fully connected layers. As a matter of fact, the model is an adapted version of the popular VGG network [19]. See Figure 4.2 for a visual representation of the model architecture.

| CNN Model | Number of layers | Model Description | Accuracy (on validation set of isolated faces dataset) |
|---|---|---|---|
| Model 1 | 5 | 3 conv + 2 fc layers | 0.5630 |
| Model 2 | 6 | 4 conv + 2 fc layers | 0.5861 |
| Model 3 | 9 | 5 conv + 4 fc layers | 0.5966 |
| Model 4 | 10 | 7 conv + 3 fc layers | 0.6158 |
| **Model 5** | **10** | **7 conv + 3 fc layers** | **0.6271** |
| Model 6 | 16 | 13 conv + 3 fc layers | 0.6365 |
| Model 7 | 15 | 13 conv + 2 fc layers | 0.6355 |

Table 4.1 - Various CNN models with their corresponding number of layers, descriptions and accuracies (on the validation set of isolated faces dataset)

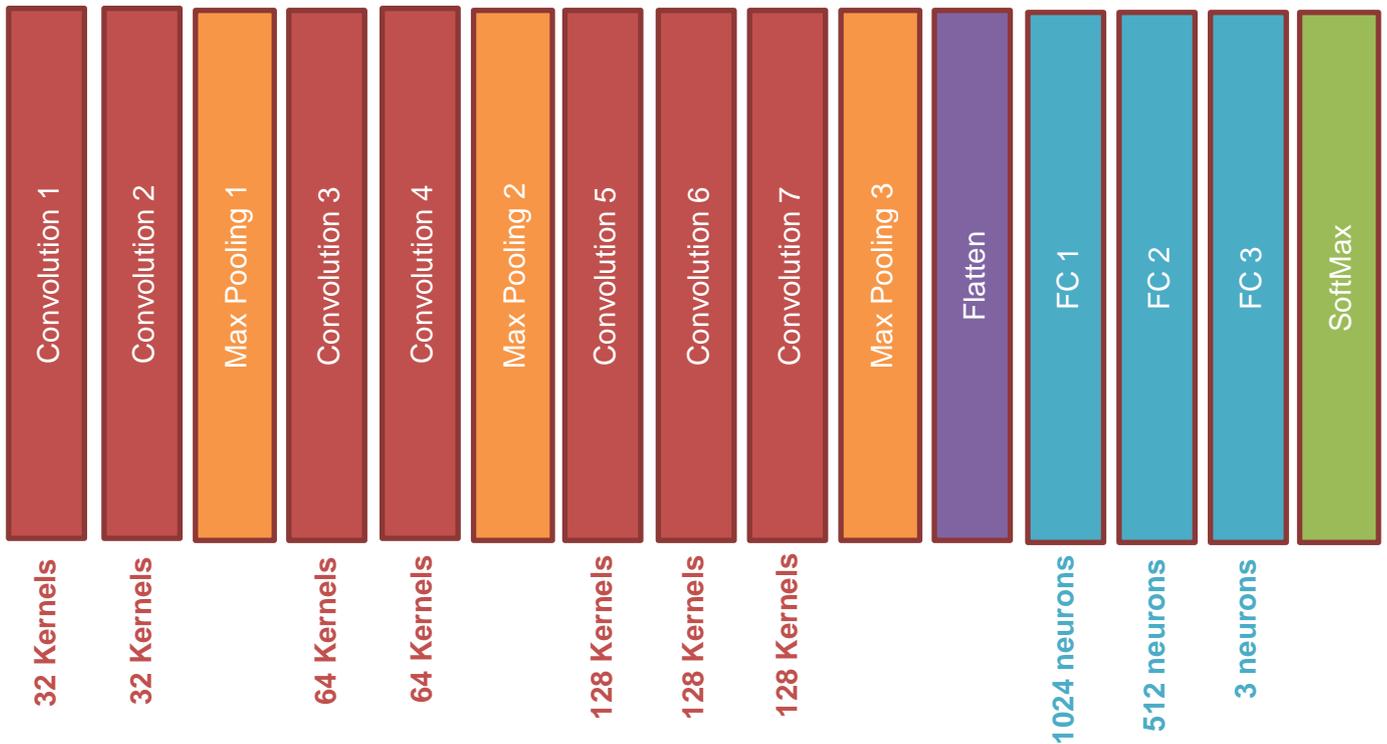

Figure 4.2 - Proposed CNN architecture consisting of 7 convolutional, 3 max-pooling and 3 fully connected layers.

#### 4.1.3.1.1 Convolutional Layer

Convolutional layers are the core building blocks of a CNN. They consist of various independent learnable filters (also referred to as kernels) which have a small perceptive field but cover the full depth of the input volume.

During the forward pass, we take each filter and slide it over the entire image, performing a matrix multiplication operation between the filter and regions of the input image thus producing an activation map for that filter. Stacking together the activation maps for all the filters along the depth dimension we can get the full output volume of the layer.

For example, Figure 4.3 consists of 2 filters W0 and W1 of size 3 x 3, stride of 2 and zero padding of 1. It shows how -4 is obtained in the output volume (highlighted green cell) by elementwise multiplying the highlighted input (blue) with filter W0 (red), summing it up, and then adding a bias b0.

There are several hyperparameters such as the kernel size, number of kernels, and stride that control the size of the output volume of the convolutional layer. In our case, we used convolutional layers with a kernel size of 3 x 3, a stride of 3 and the number of kernels as 32, 64 or 128.

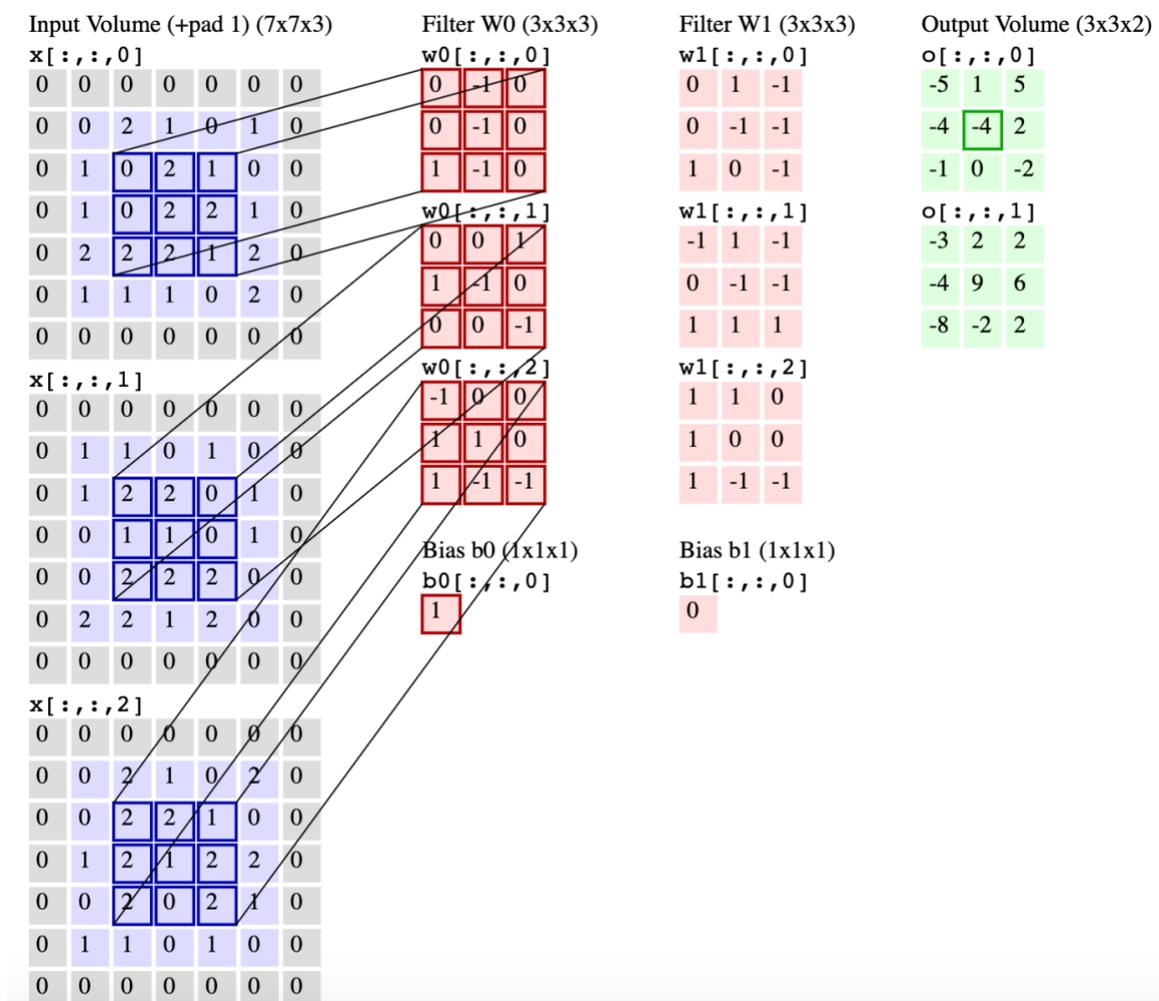

Figure 4.3 - Visualization of a convolution layer **[20]**

### 4.1.3.1.2 Activation Function

An activation function is a node that is put at the end of or in between neural networks. It takes a number as input, performs a non-linear transformation on it and then sends it to the next layer as input.

Rectified Linear Unit (ReLU) has been used as the default activation function in our CNN model. It applies the function $f(x) = \max(0, x)$ to the input values and hence converts all the negative input values to zero. This makes it computationally efficient as fewer number of neurons are activated each time.

ReLU layers work far better than sigmoid and tanh activation functions as they greatly accelerate (by a factor of 6 [21]) the convergence of stochastic gradient descent.

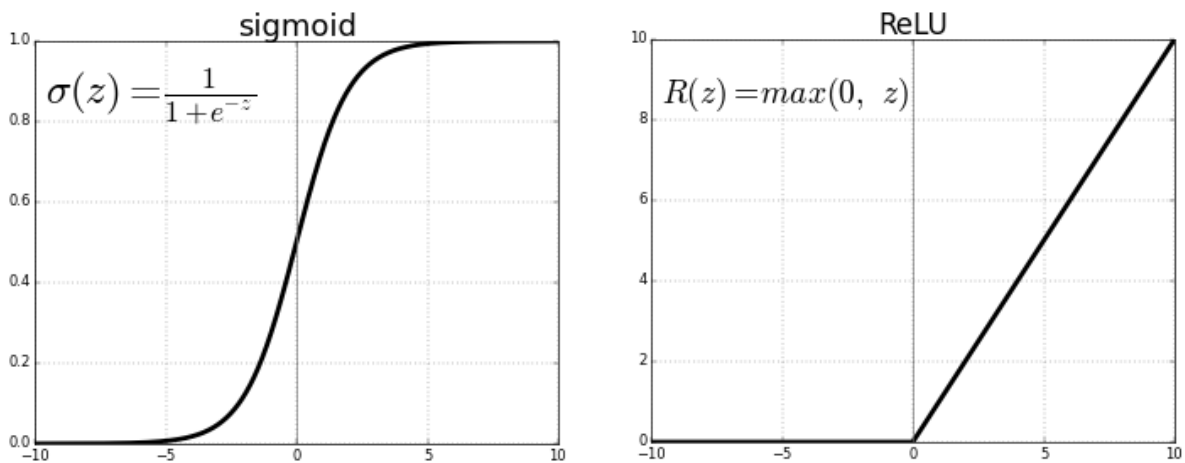

Figure 4.4 - ReLU vs sigmoid function **[22]**

### 4.1.3.1.3 Pooling Layer

Pooling is a form of non-linear down-sampling. A pooling layer is periodically inserted between consecutive convolution layers to progressively reduce the spatial size of the representation in order to reduce the number of parameters and computation in the network, and hence to control overfitting.

Max pooling which is one of the most used non-linear pooling functions has been adopted in our CNN. It takes a 2 x 2 filter with a stride of 2 and applies it to every depth slice in the input, and for each sub-region outputs the maximum number.

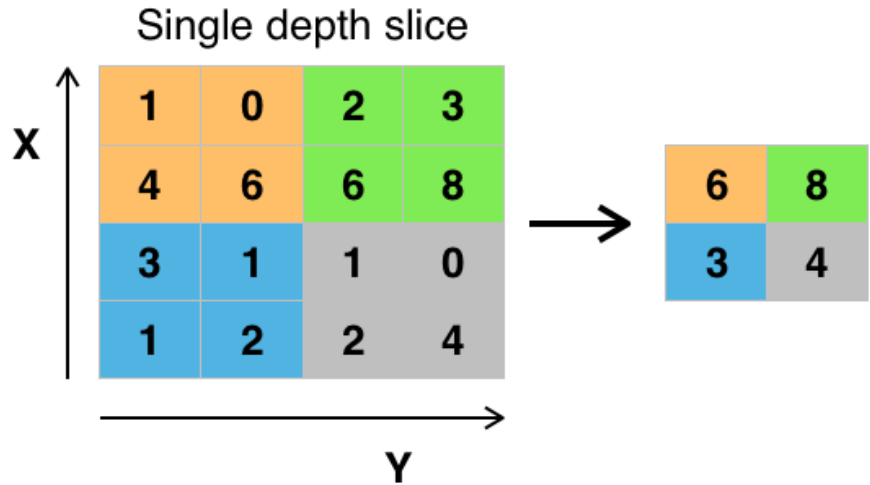

Figure 4.5 - Max pooling with a 2x2 filter and stride = 2

**4.1.3.1.4 Fully Connected Layers**

Fully connected layers are the last few layers of the neural network that take the output of the preceding layers and output an n-dimensional vector where n is the number of classes the program needs to classify from.

We tried three values - 256, 512, 1024 for the number of neurons in each fully connected (FC) layer. For the first FC layer, 1024 neurons produced the lowest validation loss. For the second FC layer, 512 neurons produced the highest validation accuracy and the lowest validation loss whereas decreasing the number of neurons to 256 and increasing to 1024 reduced the validation accuracy. Finally, for the third FC layer, 3 neurons corresponding to the three emotion classes was chosen as the default value.

**4.1.3.1.5 SoftMax Classifier**

Since this is a multi-class classification problem, i.e. we need to classify an input image as having one of the three emotion classes - positive, neutral or negative, we pass the output of the last fully-connected layer into a SoftMax Classifier instead of a Sigmoid (which is generally used for binary classification problems).

A SoftMax classifier returns a categorical probability distribution, i.e. the probabilities of each class in the range 0 to 1 wherein the sum of the probabilities of all the classes will be equal to one, and the predicted emotion class will be the one with the highest probability.

### 4.1.4 Averaging of individual predictions

To predict the overall emotion for the group image, we first took the mean of the predictions outputted by the CNN on all the faces (detected/extracted from the group image). Then, the emotion class with the highest probability was returned as the overall group emotion for the image. Both the above steps can be summarized using the equation represented below –

$$\hat{o} = argmax\left(\frac{(\sum_{k=1}^{n} \sigma(\mathbf{o}_k))}{n}\right)$$

where $n$ is the total number of faces detected in an image, **o** is a 3-d vector representing the output of the CNN (corresponding to the probabilities for the three emotion classes), σ is the SoftMax function, and $\hat{o}$ is the index of the emotion class with the highest overall probability.

## 4.2 Top-down Module

According to previous research, scene context information plays a vital role in the perception of the group affect [13] which is the reason for using scene descriptors in the top-down module. The top-down module helps compensate the misclassification errors in the bottom-up module (which centres entirely on local features) as it can better represent the global features (scene information).

In the experimental phase, Google's Vision API [17] was used to perform label detection on the images in the training set in order to obtain the scene descriptors corresponding to each image. Due to limited resources, this process was performed on just a third of the total number of images in the training set, assuming that the resulting descriptor sample will generalize to the whole training set.

Then, a histogram of scene descriptors corresponding to each of the three emotion classes was built, leading to a total of 809 unique descriptors appearing with a particular frequency in the dataset.

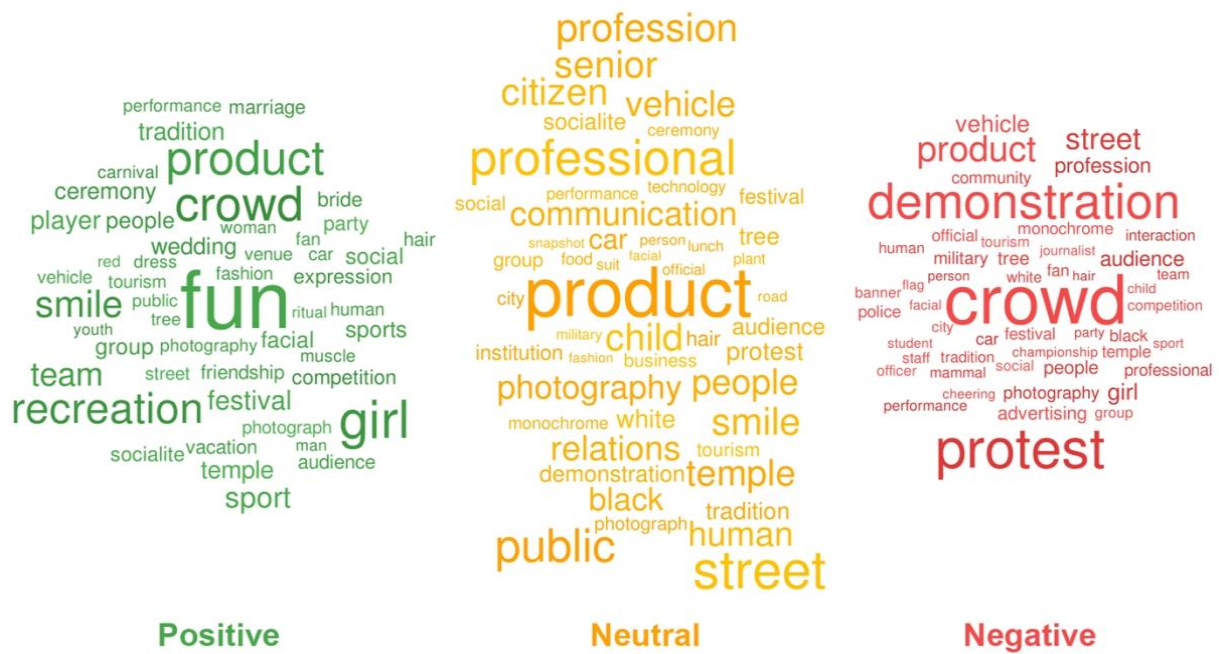

Figure 4.6 - Visualization of the top 50 most frequently occurring scene descriptors (words with larger font size => greater frequency) for each of the three emotion categories: positive (green), neutral (yellow) and negative (red).

Figure 4.6 visualizes the top 50 scene descriptors for each emotion category in the form of word clouds. It can be deduced from the image that positive words like fun, recreation, and smile frequently occur in positive images whereas negative words like protest and demonstration frequently occur in negative images.

## Bayesian Networks

Bayesian networks are a type of Probabilistic Graphical Model that represent a set of random variables and their conditional dependencies via a directed acyclic graph (DAG). They have been used previously in a wide range of applications including anomaly detection, prediction, medical diagnosis, time series prediction, and automated insights.

In our method, the Bayesian network consists of a root node, i.e., the emotion node which is a multinomial probability distribution with three possible outcomes corresponding to the three emotion classes. Each of the 809 scene descriptors obtained earlier were set as dependent nodes and represented with a Bernoulli distribution: true (present), false (absent) and an edge directed towards the emotion node.

The posterior probability for the emotion node $y$ is directly proportional to the product of its prior probability $P(y)$ and the likelihood $P(x_i|y)$ for each of the N dependent variables $x$ –

$$P(y|x_1, x_2, \ldots, x_N) = P(y) * \prod_{i=1}^{N} P(x_i|y)$$

In the equation above, we already know the prior probability $P(y)$ since we know the emotion associated with each image in the training set. The conditional probability $P(x_i|y)$ for each dependent variable can be estimated using Maximum Likelihood Estimation (MLE) [23].

For example, let $N_t^+$ be the total number of times a specific scene descriptor $x_i$ has been encountered in positive images and $N_f^+$ be the total number of times it has not been encountered. Then, the conditional probabilities of $x_i$ being associated with positive emotion can be calculated as –

$$P(x_i = true \mid y = positive) = \frac{N_t^+}{N_t^+ + N_f^+}$$

$$P(x_i = false \mid y = positive) = 1 - P(x_i = true \mid y = positive)$$

Similarly, we can also estimate the conditional probabilities of $x_i$ being associated with neutral and negative emotions respectively. These conditional probabilities can be used to calculate the Bernoulli distributions for each descriptor in order to produce the corresponding conditional probability distribution tables to initialize the Bayesian network.

In order to make predictions on an input image using the Bayesian network, the following steps are followed –

### 4.2.1 Acquiring the scene descriptors

First of all, we use Google Vision API's [17] label detection feature to obtain a list of context-specific (e.g., party, vacation, picnic, etc.) and group-specific (e.g., friends, school, military, etc.) scene descriptors specific to the input image.

### 4.2.2 Setting evidence in the Bayesian Network

In this step, the image scene descriptors obtained using the Vision API [17] in the previous step are used to set evidence for the dependent nodes defined previously in the Bayesian network.

### 4.2.3 Estimating the posterior probabilities of the Bayesian Network

In the final step, an inference is made on the Bayesian network using the Variable Elimination method [24] in order to obtain the emotion probabilities for the input image. Then, the emotion class with the highest probability is returned as the final group emotion for the image.

Essentially, we want to estimate the emotion class with the highest probability (ŷ) given a set of observed scene descriptors (obtained in 4.2.1) which can be represented mathematically as –

$$\hat{y} = argmax\ (P(y|x_1, x_2, \ldots, x_N))$$

## 4.3 Integration

Several methods like averaging, weighted mean, ensemble averaging and redirection of the output of the bottom-up module as input into the top-down module can be used to fuse together the bottom-up and top-down modules. After careful experimentation, it was found that the redirection method achieves the highest accuracy on the validation set and hence it was chosen as the default method.

In this method, an additional dependent node $x_{N+1}$ (which is a 3-category multinomial probability distribution corresponding to the three emotion classes) representing the CNN is introduced in the Bayesian Network. Like before, the conditional probability distribution table for this node needs to be calculated and this can be achieved using the confusion matrix for the CNN obtained during the validation phase.

To obtain the prediction using the full pipeline, the index of the emotion class predicted by the CNN as well as the acquired scene descriptors are used to set evidence in the Bayesian network and then just like before; an inference is made on the Bayesian network using the Variable Elimination method to obtain the final group emotion.

# Chapter 5 – Experimental Design

In order to build an efficient neural network, various factors such as network architecture and input data format need to be taken into account. This section presents a series of experiments that were carried out as well as the methodology that was followed to train the convolutional neural network (in the bottom-up module) and achieve the desired results.

## 5.1 Face Detection

Face detection plays a crucial role in the proposed approach. It is not just used in the final pipeline to detect the faces in a group image for further facial expression analysis but also for generating the dataset needed to train the individual emotion CNNs. Some of the facial detection algorithms that we tried are presented below -

### 5.1.1 Haar Cascade Face Detector in OpenCV

Face Detection using Haar feature cascade classifiers [25] is an effective machine learning based approach which was the state-of-the-art for many years since 2001 when Paul Viola and Michael Jones proposed it. In this method, a cascade function is trained on lots of images with and without faces, and then used to detect faces in other images. There have been many improvements in this approach in recent years.

Pros -
1. Almost real-time detection.
2. Detects faces on different scales.

Cons -
1. Lots of false positives.
2. Fails to work on non-aligned images and those with occlusions.

### 5.1.2 DNN Face Detector in OpenCV

OpenCV's Deep Neural Network face detector is based on the Single Shot Detector (SSD) framework [26] with a ResNet-10 Architecture [27] as the backbone. It was trained on lots of facial images from the web.

Pros -
1. Very accurate face detection.
2. Runs in real-time.
3. Detects faces on different scales.
4. Works on non-aligned images as well as those with occlusions.

Cons -
1. Ignores a lot of faces, i.e. detects a smaller number of faces compared to other algorithms.

### 5.1.3 HoG Face Detector in Dlib

Dlib's face detector is based on Histogram of Oriented Gradients (HOG) [28] features combined with a linear classifier (SVM) [29]. It was trained on 2825 images obtained from the LFW dataset [30].

Pros -
1. Fastest face detection method.
2. Works on images with small occlusions.

Cons -
1. Does not detect faces with size less than 80x80 pixels.
2. Fails to work on non-aligned facial images and those with big occlusions.

### 5.1.4 MTCNN Face Detector

Multi-Task Cascaded Convolutional Neural Network (MTCNN) [16] is a deep-learning based approach that uses three cascaded convolutional neural networks (CNNs) for fast and accurate face detection. In MTCNN, face detection and face alignment are done together in a multi-task training fashion, allowing it to detect non-aligned faces better.

Pros -
1. Works in real-time on GPU.
2. Very accurate face detection.
3. Detects faces on different scales.
4. Works for different face orientations.
5. Works on images with occlusions.

Cons -
1. Slow on CPU.

### Analysis
In order to choose the face detection algorithm ideal for this task, all four methods were tested on a total of 300 randomly selected images from the training set (100 images belonging to each emotion class). The number of faces detected by each method for the 300 images can be observed in table 5.1.

| Method | Number of faces detected |
|---|---|
| Haar Cascade Face Detector in OpenCV | 1138 |
| DNN Face Detector in OpenCV | 466 |
| HoG Face Detector in Dlib | 963 |
| MTCNN Face Detector | 1309 |

Table 5.1 - Table representing the number of faces detected by each of the four face detection methods on a set of 300 images from the training set.

In our experiments, we found MTCNN's face detector to be the most accurate one on the EmotiW dataset followed by Dlib. However, MTCNN's face detector was quite slow on CPU. DNN Face Detector in OpenCV detected a small number of faces whereas Haar Cascade Face Detector in OpenCV detected a good number of faces but yielded many false positives.

Hence, in the final pipeline, we decided to use Dlib's HoG based face detector. However, for some images, it was not able to detect any faces. Thus, to obtain ample amount of isolated face training data, we decided to use both Dlib, and MTCNN face detection methods sequentially, i.e. first we run Dlib's face detector, and if it does not detect any faces, then we try the MTCNN approach.

## 5.2 Data Pre-processing

Data Pre-processing is a crucial step that transforms the raw data before it is fed into a neural network and helps significantly improve the classification performance of the neural network. Furthermore, it helps speed up the training process of the neural network. Some of the pre-processing techniques tried are presented below -

### 5.2.1 Cropping

After performing face detection on the group image [See Section 5.1], OpenCV was used to crop and save the individual detected faces using the coordinates provided (by the face detection algorithm).

### 5.2.2 Scaling

After ensuring that all the individual detected faces had been cropped and saved, it was time to scale each facial image (obtained in the previous step) appropriately as it is computationally expensive to train a CNN on large images.

We decided to have images of size 64x64 pixels. The original width and height ratio for each image was calculated, and then OpenCV's resize function was used to rescale each image to 64x64 pixels while maintaining the original size ratios.

### 5.2.3 Normalization

Normalization is a crucial image pre-processing technique that ensures that each pixel in the picture has similar intensity values, which in turn makes the convergence faster while training the neural network. So, each image was divided by 255 to ensure that each feature was in the range [0, 1].

### 5.2.4 Facial Alignment

Face alignment is a computer vision technique for identifying the geometric structure of faces in images and obtaining a canonical alignment of the faces using image transformation.

Dlib's facial alignment module was used to determine the orientation of the faces and obtain the skewed and transformed facial images with all the facial features aligned appropriately.

### Analysis

From the pre-processing techniques mentioned above, cropping and normalization are absolutely necessary for the individual emotion CNN to achieve satisfactory performance. Adding in scaling and facial alignment to the mix helped improve the classification accuracy of the individual emotion CNN by 5%.

## 5.3 Isolated Faces Dataset

For each image in the training set (of GAF 3.0 dataset), we isolated the faces using facial detection algorithms as defined in section 5.1. Then, we performed a series of pre-processing operations as explained in section 5.2 in order to transform the raw image data.

These two steps result in a new dataset containing 32k isolated facial images of size 64 x 64 in the training set and 15k in the validation set, that was used later on to train our individual facial emotion CNN models in the bottom-up module. This new dataset will be referred to as the "isolated faces dataset" from here on.

## 5.4 Individual Facial Emotion CNN Experiments

Several state-of-the-art CNN models (to be used in the bottom-up module) were implemented and trained on the individual facial images extracted from the training set [refer to Section 5.3 for more information]. This section presents the series of experiments that we carried out to train the CNN models efficiently in order to minimize the validation loss and maximize the validation accuracy.

### 5.4.1 Hyperparameter Optimization

A hyperparameter is a parameter whose value is used to control the training process of the neural network. Hyperparameter optimization is the process of choosing a set of optimal hyperparameters for a learning algorithm in order to improve the quality of results that we are trying to obtain.

There are various methods for optimizing hyperparameters such as grid search, random search, and Bayesian optimization. For this project, the random search technique has been used wherein random combinations of a range of values for each hyperparameter have been used to find the best possible combination. The effectiveness of the random search technique for hyperparameter optimization has been demonstrated in [31].

Table 5.2 depicts all the hyperparameters used by the proposed model with their corresponding values.

| Hyperparameter | Value | Description |
| --- | --- | --- |
| Batch size | 128 | The number of images (from the training set) to be propagated through the network at once. |
| Number of epochs | 20 | The number of complete passes through the training set. |
| Optimizer | Adam | Optimization algorithm. |
| Learning rate | 0.001 | Determines how fast the weights of the network are updated during training. |
| FC1 neurons | 1024 | Number of neurons in the first fully connected layer. |
| FC2 neurons | 512 | Number of neurons in the second fully connected layer. |
| Dropout | 0.5 | Dropout rate between fully connected layers. |
| Convolution kernel size | 3x3 | Size of the kernel in each convolution layer. |
| MaxPooling kernel size | 2x2 | Size of the kernel in each MaxPooling layer. |
| MaxPooling strides | 2 | Kernel stride in each MaxPooling layer. |

Table 5.2 - Hyperparameters used by the proposed model with their corresponding values.

### 5.4.2 Reduce Overfitting

One of the problems that occur during the training of neural networks is overfitting. Overfitting happens when the weights of a model are so tuned to the training data that it negatively impacts its performance on new data.

Figures 5.1 and 5.2 depict the learning curves of the network trained without implementing any overfitting prevention techniques. It can be inferred from the graphs that there is a steep increase in the validation loss (referred to as test loss in Figure 5.1) after three epochs and a 100% training accuracy after seven epochs, even though the validation accuracy (referred to as test accuracy in the Figure 5.2) is quite steady. These factors indicate that the network is clearly overfitting.

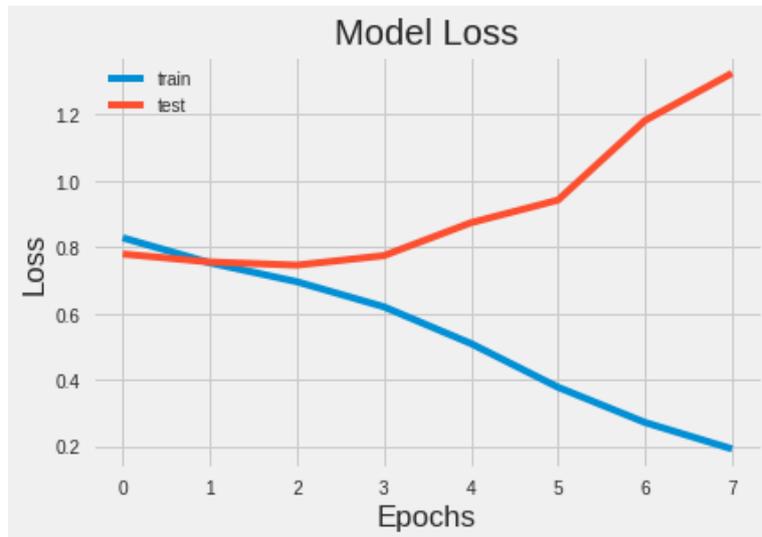

Figure 5.1 - Loss vs number of epochs learning curve of the model

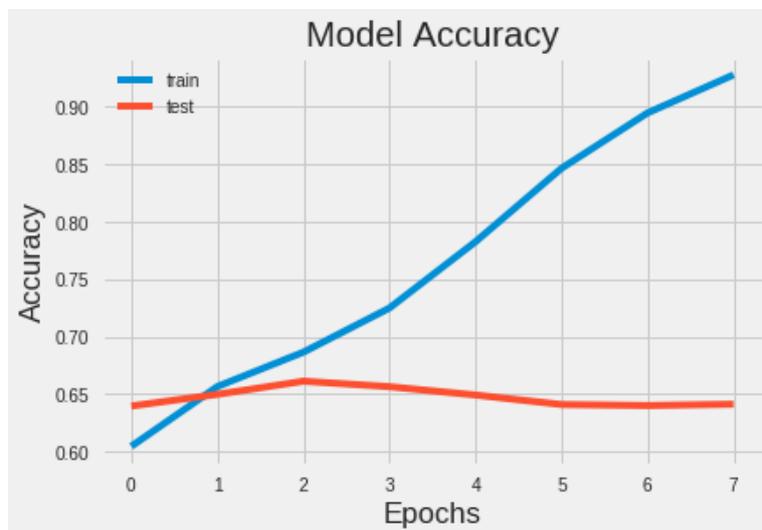

Figure 5.2 - Accuracy vs number of epochs learning curve of the model

Below we present some techniques that were employed to reduce overfitting in the CNN model.

#### 5.4.2.1 Dropout

The basic idea behind this technique is to "drop out" a random set of activations between two fully connected layers by setting them to zero in order to leave a reduced network. This layer is only used during the training phase, making sure that the network doesn't get too "fitted" to the training data so that it can better generalize to new data. Thus, it not only helps overcome the overfitting problem but also significantly improves training speed.

In our baseline CNN model, a dropout of 0.5 has been used, i.e., half of the activations between consecutive fully connected layers have been cancelled. After careful experimentation, it was found that introducing a dropout layer helped increase the validation accuracy by 6%.

#### 5.4.2.2 Image Augmentation

Deep neural networks require large amounts of training data in order to achieve good results. Image augmentation is a method of artificially creating more training images using different image processing techniques to boost the performance of the CNN.

This method works very well for relatively small datasets like our isolated faces dataset consisting of just 32k sample images. A combination of three image processing operations was used to achieve image augmentation in our project –

1. Random image rotation in the range [0, 40]
2. Random zoom in the range [0, 0.1]
3. Random horizontal flips

Image augmentation was performed right after data pre-processing [please refer to Section 5.2] during the training phase only, and it was performed on each mini-batch of tensor image data separately (in real-time) allowing the network to see different images at each epoch. Using image augmentation helped increase the validation accuracy of the model by 3.5%.

#### 5.4.2.3 Regularization

Early stopping regularization method was used to prevent overfitting further. It allowed us to specify an arbitrarily large number of training epochs and stop training once the validation loss of the model had stopped improving on a holdout validation dataset.

### 5.4.3 Training

The CNN has been trained on 32k images of size 64 x 64 from the isolated faces dataset (created in Section 5.3). During the training process, we randomly selected a batch of 128 images from the training set and performed gradient descent for 20 epochs. The training process took approximately 10 minutes on Google Colab [32].

### 5.4.3.1 Categorical Cross entropy Loss

Cross-entropy loss (also known as log loss) is a loss function that measures the performance of a classification model whose output is a probability in the range [0, 1].

For this project, we have used the categorical cross-entropy loss (also known as SoftMax loss), which is basically a SoftMax activation plus a Cross-Entropy loss allowing us to train our CNN to output a probability over the three emotion classes for each image.

### 5.4.3.2 Adam optimizer

As an optimizer, we use Adam [33], which has been chosen after a preliminary comparison between other methods such as Adagrad and RMSProp. Adam is an adaptive learning rate method, meaning that it computes individual learning rates for different parameters using estimations of first and second moments of the gradients. It showed substantial performance gains in the model training speed.

**Require:** $\alpha$: Stepsize
**Require:** $\beta_1, \beta_2 \in [0, 1)$: Exponential decay rates for the moment estimates
**Require:** $f(\theta)$: Stochastic objective function with parameters $\theta$
**Require:** $\theta_0$: Initial parameter vector
$\quad m_0 \leftarrow 0$ (Initialize 1st moment vector)
$\quad v_0 \leftarrow 0$ (Initialize 2nd moment vector)
$\quad t \leftarrow 0$ (Initialize timestep)
$\quad$**while** $\theta_t$ not converged **do**
$\quad\quad t \leftarrow t + 1$
$\quad\quad g_t \leftarrow \nabla_\theta f_t(\theta_{t-1})$ (Get gradients w.r.t. stochastic objective at timestep $t$)
$\quad\quad m_t \leftarrow \beta_1 \cdot m_{t-1} + (1 - \beta_1) \cdot g_t$ (Update biased first moment estimate)
$\quad\quad v_t \leftarrow \beta_2 \cdot v_{t-1} + (1 - \beta_2) \cdot g_t^2$ (Update biased second raw moment estimate)
$\quad\quad \widehat{m}_t \leftarrow m_t / (1 - \beta_1^t)$ (Compute bias-corrected first moment estimate)
$\quad\quad \widehat{v}_t \leftarrow v_t / (1 - \beta_2^t)$ (Compute bias-corrected second raw moment estimate)
$\quad\quad \theta_t \leftarrow \theta_{t-1} - \alpha \cdot \widehat{m}_t / (\sqrt{\widehat{v}_t} + \epsilon)$ (Update parameters)
$\quad$**end while**
$\quad$**return** $\theta_t$ (Resulting parameters)

Figure 5.3 - Adam optimization algorithm (as presented in the original paper)

# Chapter 6 – Evaluation

In this section, we evaluate our method on the validation set of the GAF 3.0 dataset. Evaluation of a learning algorithm means measuring how close the predicted emotion labels are to the real labels when tested on some previously unseen data.

Classification accuracy has been chosen as the metric for evaluating the performance of the proposed approaches. The higher it is, the better the model performs. Additionally, confusion matrices have been produced to visualize the performance of each method.

## 6.1 Evaluation of the bottom-up module

Table 6.1 shows the results of the top 4 best performing individual emotion CNN models as well as an ensemble of those models on the GAF 3.0 validation set. It can be observed from the table that the ensemble model significantly outperforms the results obtained by the individual emotion CNN models. Hence, the ensemble model has been chosen as the default CNN model for the bottom-up module.

To further evaluate the ensemble model and to better understand how classification accuracies for each emotion class compare to each other, a confusion matrix on the images of the validation set was created.

Face detection is one of the crucial steps of this module. The confusion matrix in Figure 6.1 shows that quite a few photos have been classified as "None." Here, "None" implies that no face has been detected in those images due to factors like occlusions, image quality, variance of actors, head and body pose variations, etc. As a result, no prediction could be made. This is one of the drawbacks of the bottom-up module as it is dependent on the robustness of the face detection algorithms. It can also be observed that the model performs remarkably well on positive images because this emotion has easily distinguishable features compared to the other emotions. Furthermore, the neutral emotion has often been misclassified as negative and vice versa as both these emotions produce somewhat similar features.

| CNN Model | Accuracy (on validation set of GAF 3.0) |
|---|---|
| Model 6 [See Appendix A] (1) | 0.5858 |
| Model 7 [See Appendix A] (2) | 0.5757 |
| Model 5 (3) | 0.5959 |
| Model 5 with 2048 neurons in FC1 (4) | 0.5858 |
| **Ensemble Model (1+2+3+4)** | **0.6426** |

Table 6.1 - Accuracy of various individual emotion CNN models in the bottom-up module on the validation set.

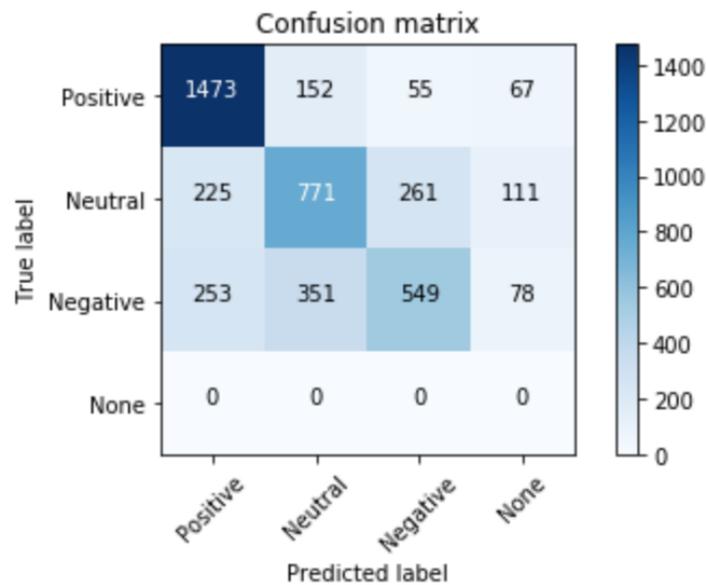

Figure 6.1 - Confusion Matrix for the ensemble model on the validation set.

## 6.2 Evaluation of the top-down module

Table 6.2 shows the results of the Bayesian network in the bottom-up module on the GAF 3.0 validation set. It can be observed that the top-down module achieves a much lower accuracy than the bottom-up module.

Figure 6.2 shows the confusion matrix of the Bayesian network on the images of the validation set. It can be noted that quite a few positive as well as negative images have been classified as neutral. This is because the neutral emotion is much more difficult to define.

| Model | Accuracy (on validation set) |
|---|---|
| **Bayesian Only** | **0.6104** |

Table 6.2 - Accuracy of the Bayesian Network in the top-down module on the validation set.

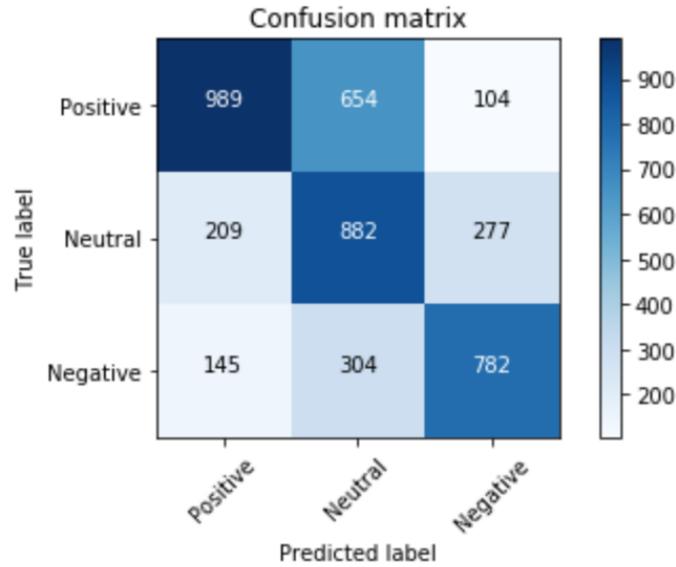
Figure 6.2 - Confusion Matrix for the Bayesian network on the validation set.

## 6.3 Evaluation of the full pipeline

Table 6.3 shows the results of the full pipeline (Bayesian network + CNN) on the validation set. It can be noted that the fusion of the bottom-up and top-down modules achieves remarkably higher accuracy than the isolated modules which is in accordance with a previous paper that demonstrated how an ensemble could perform better when used in emotion recognition systems [34].

The final pipeline achieves 65.27% accuracy on the validation set of GAF 3.0 dataset which is similar to the baseline results [5] obtained by the organizers of the competition who used a very deep Inception v3 architecture. However, we achieved comparable results with much less resources and computing power.

Figure 6.3 shows the confusion matrix of the full pipeline on the images of the validation set.

| Model | Accuracy (on validation set) |
|---|---|
| Bayesian + CNN | 0.6527 |

Table 6.3 - Accuracy of the full pipeline (CNN + Bayesian Network) on the validation set.

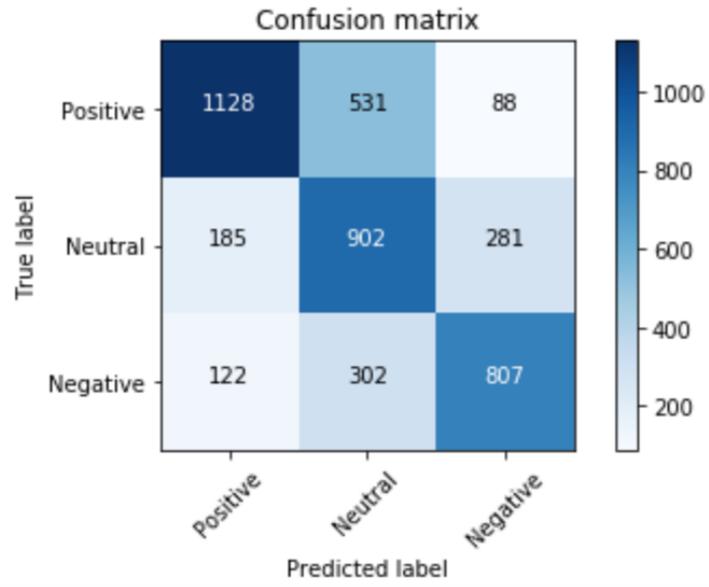
Figure 6.3 - Confusion Matrix for the full pipeline on the validation set.

# Chapter 7 – Tools and Technologies

1. **Language**: The project has mainly been coded in Python 3.7 [35] due to its extensive selection of libraries and frameworks, and ease of use.

2. **Deep Learning Library/Framework**: The CNN models have been trained using Keras [36]. Keras is a high-level API on top of TensorFlow [37] which is modular, Pythonic, and super easy to use, allowing the user to quickly build and test a neural network with minimal amount of code.

3. **Web Application Framework**: The web application has been developed in Flask [38] which is a Python based web development framework.

4. **Mobile Application**: The android app has been coded in Java using Android Studio [39].

5. **Cloud Service:** Training deep neural networks for image classification tasks require huge amount of processing power in order to perform fast computations. As a result, GPU capabilities have been explored.

   Google's free cloud service called Colab [32] which is basically a jupyter notebook on cloud has been used to train the CNN models. Machine configuration -
   - 12 GB RAM
   - Nvidia Tesla K80 GPU

6. **Libraries**
   - **Visualisation**: Matplotlib [40]
   - **Image Manipulation**: OpenCV [41] and Dlib [42]
   - **Scientific Computing**: NumPy [43], Pandas [44] and sklearn [45]
   - **Probabilistic Modelling**: pgmpy [46]
   - **Label Detection:** Google Vision API [17]

# Chapter 8 - Deliverables

Two deliverables were developed as part of this project - a Progressive Web Application and an accompanying Android app with a simple and intuitive user interface, allowing users to test out the system with their own pictures.

The apps can reliably recognise the overall group emotion of images uploaded from the local storage (of a device) or taken immediately from the camera of a smartphone/tablet.

## 8.1 Web Application

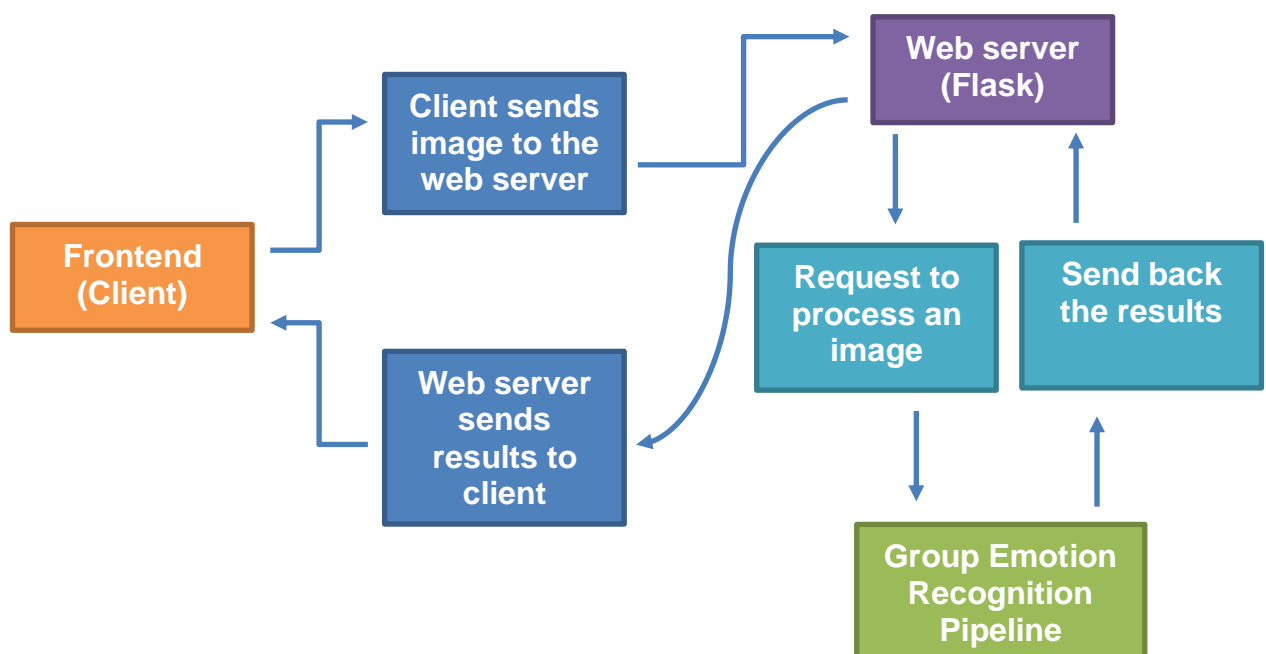

Figure 8.1 – Data flow diagram for the web application

### 8.1.1 Backend

The backend consists of two parts: a web server and the group emotion recognition module.

#### 8.1.1.1 Web Server

The role of the web server is to connect the frontend (client) with the group emotion recognition pipeline. The web server is implemented in Python 3.7, using the Flask microframework. Communication between the web server and the client is done over HTTP, and the requests and responses are all in JSON format.

The web server gets the image uploaded by the user, resizes it to 1000x1000 (to avoid processing delays of large images) and saves it to the local directory of the application. It then calls the group emotion recognition module to get the CNN only predictions, Bayesian network only predictions as well as CNN + Bayesian Network predictions for the uploaded image.

### 8.1.1.2 Group Emotion Recognition Pipeline

The group emotion recognition pipeline is implemented as a module in Python 3.7. It calls the image processing module to process the input image as well as do label detection on it, and it calls the CNN and Bayesian network modules to get the respective emotion predictions for the image. It then returns the received predictions in JSON format to the web server so that it could be presented to the user.

**a) Image Processing Module**
The image processing module is composed of various facial detection algorithms that return the coordinates of the detected faces in the image.

It also crops the detected faces (using the coordinates specified by the facial detection algorithms), rescales them to 64x64, normalizes them and aligns them if not properly aligned and then finally saves them to the local storage to be used by the CNN.

It is also responsible for calling the Google Vision API to do label detection on the input image to obtain the scene descriptors to be used by the Bayesian network.

**b) CNN Module**
Deep learning models can take hours to train, so it is essential to save the model to disk to make predictions using it in the future.

Once the CNN models have been trained and evaluated, the structure of the best model is described and saved to a JSON file, and the model weights are saved to an HDF5 file (which is ideal for storing multi-dimensional arrays of numbers).

CNN module is responsible for loading the pre-trained CNN model from the saved files and compiling it before use. It then uses the loaded model to predict the emotions for each detected (and pre-processed) face in the image and calculates the mean predictions for all the faces in that image to estimate the overall emotion for that image.

**c) Bayesian Network Module**
Bayesian network module is responsible for loading the Bayesian network model using the label histogram, setting evidence using the detected scene descriptors and performing inference on the Bayesian network to get the global emotion predictions for the image.

It is also responsible for integrating the top-down and bottom-up modules by redirecting the output of the CNN as a dependent node in the Bayesian network to get the final prediction for the image.

## 8.1.2 Frontend

Frontend components have been developed using Javascript, HTML, and CSS that communicate with the backend using JQuery. The frontend is a single page website where a user can upload an image and view the predicted emotion probabilities in the form of creative visualisations.

It is also a fully functional Progressive Web App (PWA) which means that it is fully responsive and can work on any device, taking advantage of the native features available on the user's device and browser. It is also fast, reliable and provides app like interactions to the user.

### 8.1.2.1 Interface

The app provides a simple and intuitive interface. Its design has been inspired by Google's material design, giving it a contemporary look and feel and at the same time providing a more engaging user experience with perfect usability.

It uses progressive disclosure to avoid information overload by implementing a hierarchy of operations moving through simple selectable steps, thus making it faster and easier for the user to understand.

First, the user is asked to either upload an image from their device or choose one from the predefined collection. Once the user selects/uploads a picture, they are taken to the processing section where they are shown the image they selected and asked to press the "Process image" button to process that image and get the emotion predictions for it. After the image has been processed, they are taken to the predictions section where the overall group emotion predictions are presented to them in the form of animated emojis. They are also presented with all the individual detected faces from that image with their corresponding CNN predictions as well as the group predictions made by 1) CNN only 2) Bayesian network only 3) CNN + Bayesian network.

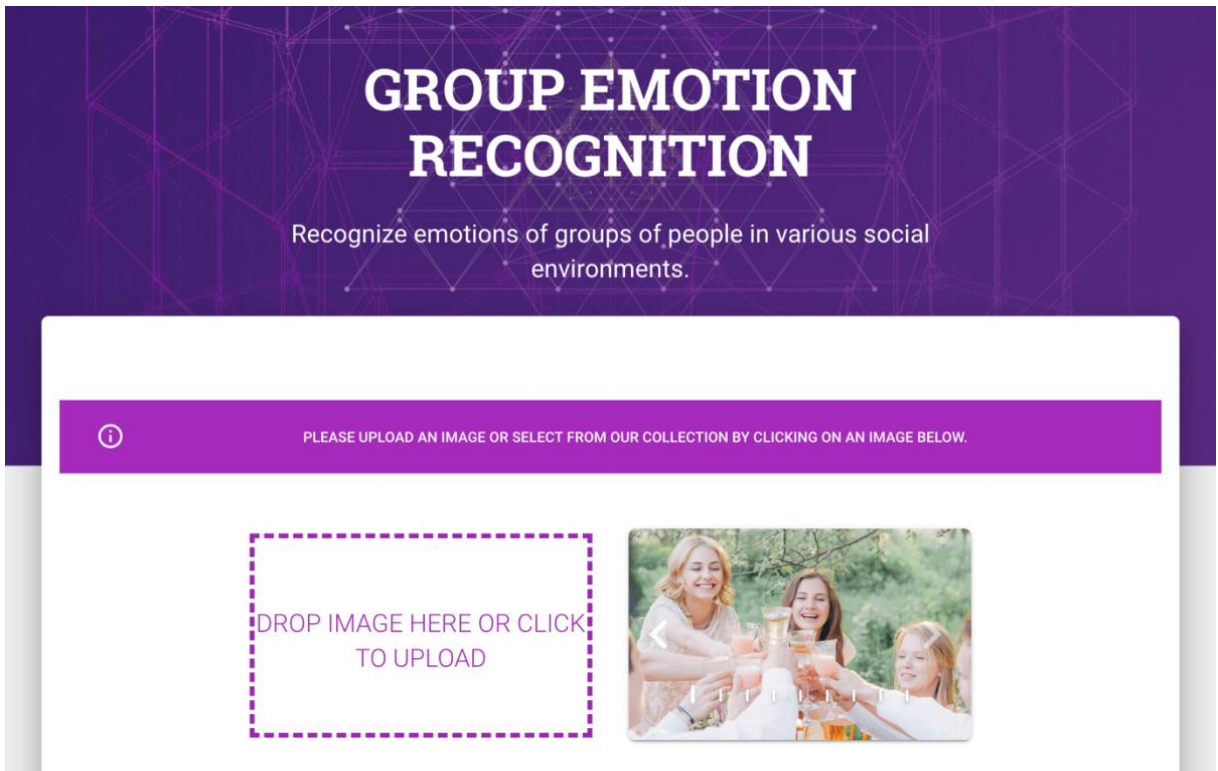

Figure 8.2 - Intro section of the web app where the user is asked to upload their own image or select one from the collection

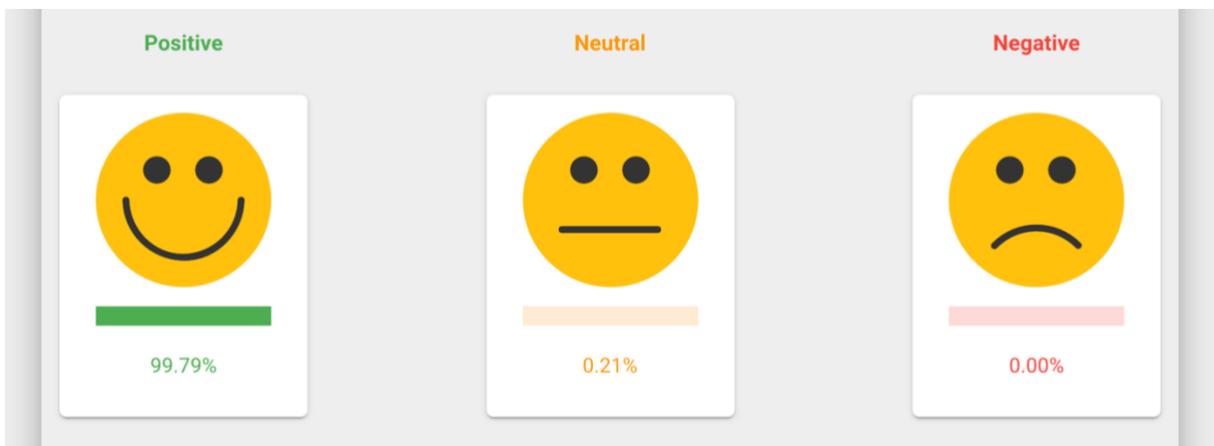

Figure 8.3 - Overall group emotion prediction for the image represented in the form of emojis

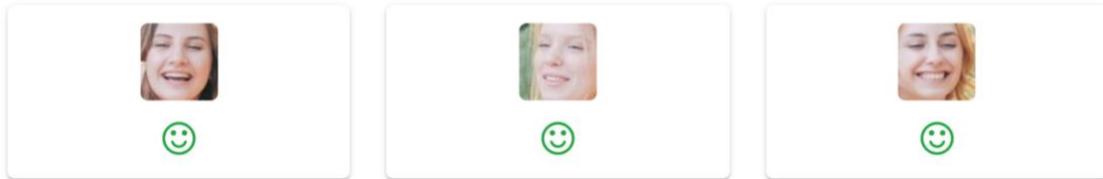

Figure 8.4 - CNN predictions for the individual faces detected in the image (represented in the form of emojis)

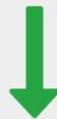

Figure 8.5 - CNN, Bayesian Network and CNN + Bayesian Network Group predictions for the image (represented in the form of tables)

## 8.2 Android Application

The Android app is written in Java using Android Studio. It is basically a hybrid app that uses a WebView to create a wrapper around the web application along with Android's native elements, allowing the user to access their device's camera to take pictures. This helps make the app maintainable as just the code for the web application needs to be updated when a change is required instead of updating both the apps separately.

Since, the android app is just a wrapper around the web app, it provides a similar user interface. What differentiates the android app from the web app is its ability to allow the user to use either camera or stored images as the input image source (See Figure 8.6). Since all the processing is done in the cloud (on the web server), the app can reliably recognise the group emotion regardless of the computing power available on the user's device thus making the app very accessible.

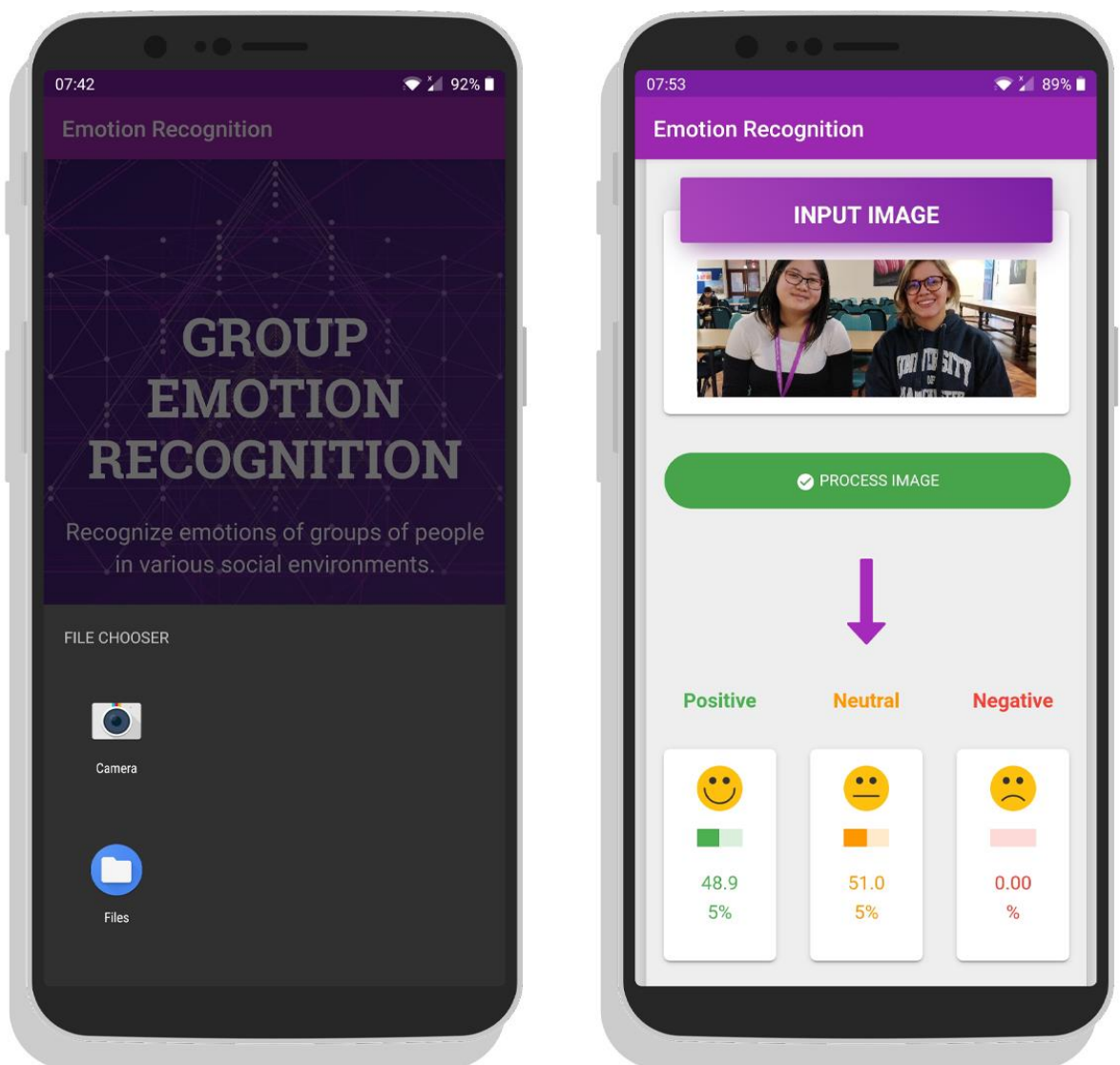

Figure 8.6 - (left) Image selection in android app giving 2 options - camera and gallery, (right) Resulting overall group emotion represented using emojis.

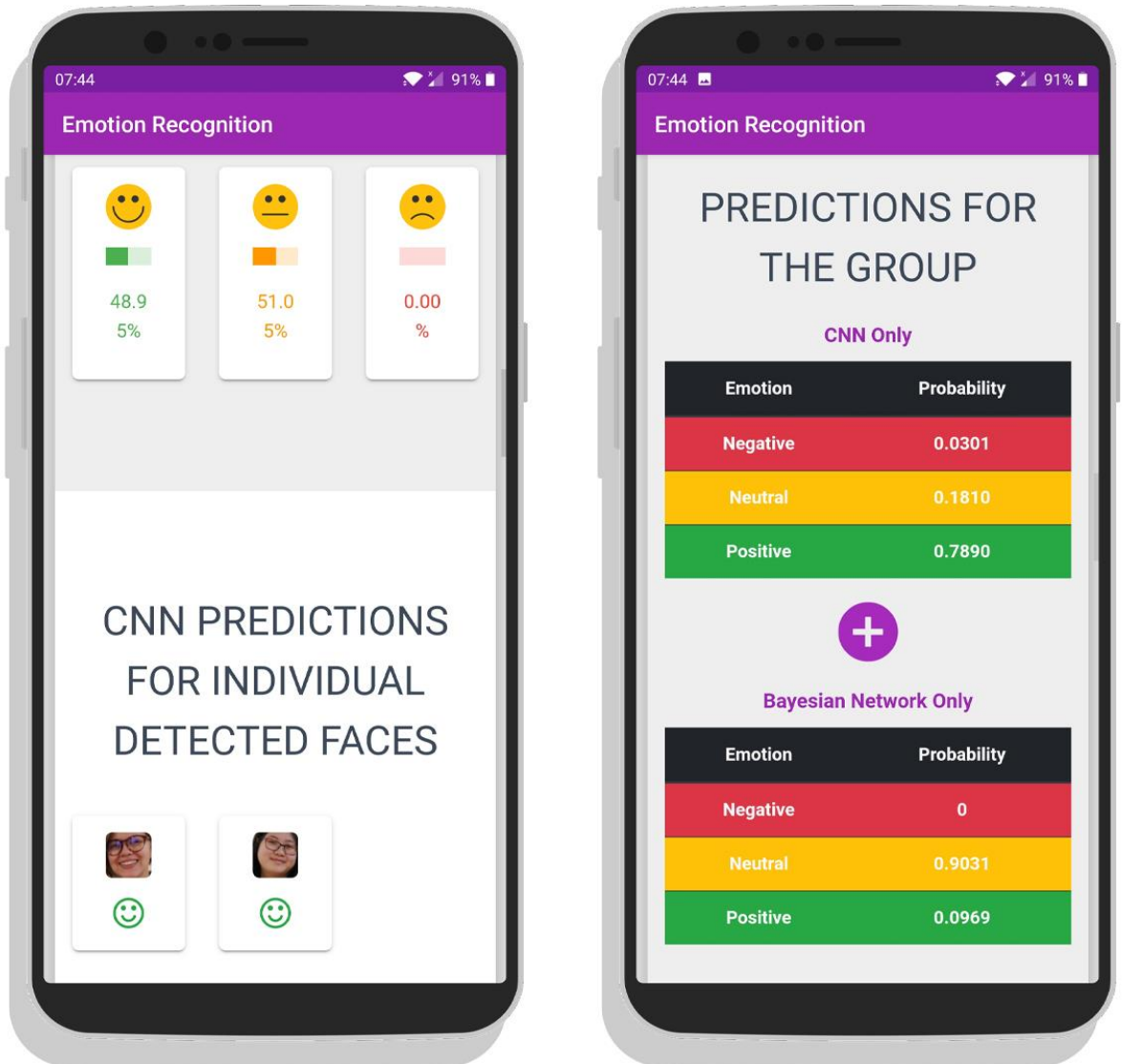

Figure 8.7 - (left) Resulting group emotion predictions as well as CNN predictions for individual detected faces, (right) CNN only and Bayesian network only group emotion predictions.

# Chapter 9 - Conclusion

In this report, we examined the use of a **hybrid machine learning system that incorporates deep neural networks and Bayesian classifiers** for group emotion recognition in the wild.

- Convolutional Neural Networks (CNNs) work from bottom to top, analysing facial expressions expressed by individual faces extracted from the image.
- The Bayesian network works from top to bottom, inferring the global emotion for the image, by integrating the visual features of the contents of the image obtained through a scene descriptor.
- In the final pipeline, the group emotion category predicted by an ensemble of CNNs in the bottom-up module is redirected to the Bayesian Network in the top-down module as a dependent node and an overall prediction for the image is obtained.

A considerable amount of research and experimentation in machine learning and deep learning methods has been carried out.

- Several advanced face detection algorithms have been implemented and tested in order to detect the maximum number of individual faces in each group image.
- Various state-of-the-art deep CNN models have been successfully trained and tested.
- Numerous techniques like dropout, data augmentation, regularisation and early stopping that help reduce overfitting have been utilised.
- Furthermore, multiple image pre-processing techniques like cropping, scaling, alignment, and normalisation have been used resulting in significant accuracy gain.
- Additionally, hyperparameter optimisation has been carried out to accelerate the learning process of the CNNs and improve the quality of the results.

Experimental results show that the stated system achieves **65.27% accuracy** on the validation set of GAF 3.0 dataset (released for the Emotion Recognition in the Wild Challenge 2018 (EmotiW)) which is similar to the baseline results [5] obtained by the organisers of the competition (who used a very deep Inception v3 architecture).

As an outcome of this project, a **Progressive Web Application (PWA)** and an accompanying **Android app** with a simple and intuitive user interface are presented, allowing users to test out the system with their own pictures.

- The apps can reliably recognise overall group emotion from images uploaded from the local storage (of a device) or taken immediately from the camera of a smartphone/tablet.
- The Android app uses a WebView to create a wrapper around the PWA along with Android's native elements, allowing a user to access their device's hardware to take pictures. This helps make the app maintainable as just the code for the web application needs to be updated when a change is required instead of updating both the apps separately.

Overall, I think this project has been a success, as it was a reasonable attempt at a real-world problem, i.e. group emotion recognition. Even though we weren't quite able to achieve state-of-the-art accuracy for the given task, the project manages to make a meaningful contribution to the ever-growing field of computer vision.

## 9.1 Future Work

In future work, better methods for combining the output of CNNs for each detected face should be explored. For example, the overall emotion for the image can be calculated using a weighted average based on the size of each face in the image [8].

Different methods for ensembling the CNNs in the bottom-up module like weighted average and stacked generalisation ensemble should be examined to reduce variance and improve the performance further.

Furthermore, in our method, we assumed all individual faces in each image (belonging to a particular class) in the training set to be belonging to the same class. For example, if a group image in the training set is labelled as positive, then all the individual faces detected in that image are also assumed as positive. So, training data (32k isolated face images belonging to the 3 classes) for the CNNs is not clean. Classification accuracy for the CNNs can be significantly improved if the dataset is cleaned manually/algorithmically.

Also, due to limited resources, scene descriptors for just a third of the images in the training set were specified using the Google Vision API [17]. So, the weights for the nodes in the Bayesian Network are not entirely accurate. Classification accuracy for the Bayesian Network can be improved if scene descriptors for all the images in the training set are specified or if a better label detection algorithm is used.

Moreover, in the bottom-up module, a large-margin SoftMax loss for discriminative learning could be utilised, and two separate CNNs could be trained - one on aligned faces and the other on non-aligned faces. The scores of both the CNNs for each face in the image can then be fused to predict the overall emotion for the group. This approach showed major improvement in [47].

# Appendix A

# Top Performing CNN Models

### 1. Model 6

Model 6 is a special type of model since it doesn't need to be trained from scratch. Here, we use an approach called transfer learning wherein a pre-trained network (VGG-16 [19] in this case) is used as the starting point for a different computer vision task (facial emotion classification in this case), resulting in a faster training process.

So, we start with the VGG-16 model which has already been trained on the ImageNet dataset which is a dataset of over 14 million images belonging to 1000 classes. Then we truncate its last layer (SoftMax layer) and replace it with our own SoftMax layer that is relevant to our problem. Finally, we fine-tune the resulting network on our isolated faces dataset so that it could be used to classify the emotions of facial images.

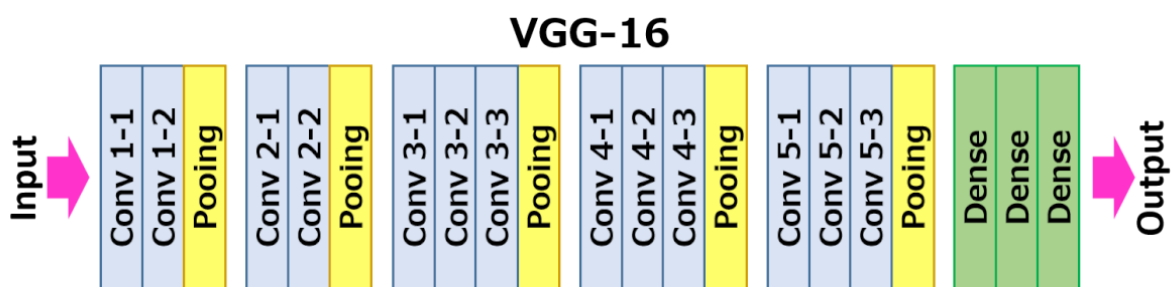

Figure A.0.1 - Architecture of VGG-16.

### 2. Model 7

Model 7 is pretty similar to Model 6; the only difference being the pre-trained network used. Here, the VGG-Face model [48] was used to perform transfer learning instead of VGG-16.

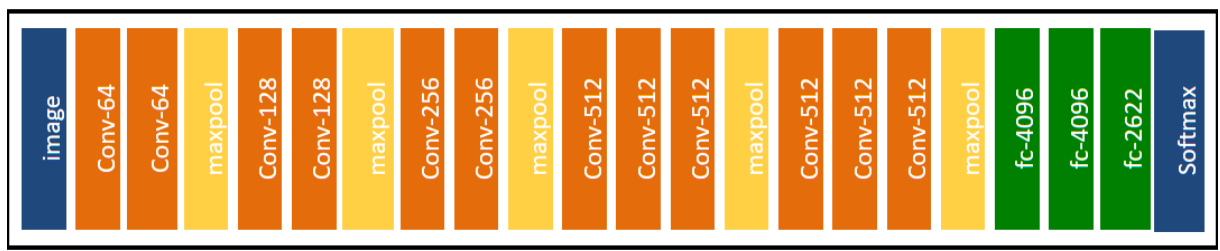

Figure A.0.2 - Architecture of VGG-Face.